# Shared Model of Sense-making for Human-Machine Collaboration

**Gheorghe Tecuci, Dorin Marcu, Louis Kaiser, Mihai Boicu**

Learning Agents Center, School of Computing, George Mason University, Fairfax, VA 22030, USA
{ tecuci; dmarcu; lkaiser4; mboicu }@gmu.edu

**Abstract**

We present a model of sense-making that greatly facilitates the collaboration between an intelligent analyst and a knowledge-based agent. It is a general model grounded in the science of evidence and the scientific method of hypothesis generation and testing, where sense-making hypotheses that explain an observation are generated, relevant evidence is then discovered, and the hypotheses are tested based on the discovered evidence. We illustrate how the model enables an analyst to directly instruct the agent to understand situations involving the possible production of weapons (e.g., chemical warfare agents) and how the agent becomes increasingly more competent in understanding other situations from that domain (e.g., possible production of centrifuge-enriched uranium or of stealth fighter aircraft).

## Introduction

Sense-making is the intelligence analysis process of situation understanding, prediction of the behavior and intent of the entities of interest, and identifying the threats as early as possible, in the context of a dynamic world, based on data that is sparse, noisy, and uncertain (Moore, 2011).

The prevailing approach to sense-making in intelligence analysis is the holistic approach where the analysts, after reviewing large amounts of information and performing the reasoning in their heads, reach a conclusion (Marrin, 2011).

A complementary approach uses very simple structured analytic techniques, such as those described by Heuer and Pherson (2011), that provide general guidelines for hypothesis generation and testing. Most of the time sense-making is the result of shallow arguments using the Toulmin intuitive model (Toulmin 1963; van Gelder, 2007), where each claim is backed by evidence. There is no systematic process to determine the probabilities of hypotheses based on the available evidence (Pherson and Pherson, 2021). More advanced methods build Bayesian probabilistic inference networks using analytical tools, such as Netica (2019), but modeling a situation with a Bayesian network is a very complex task for an intelligence analyst.

This paper presents a more advanced system for sense-making in intelligence analysis, the *Multi-Agent System for Sensemaking through Hypothesis Generation and Analysis* (MASH). MASH builds on a series of analytical tools that includes Disciple-LTA (Tecuci et al., 2008; Schum et al., 2009), TIACRITIS (Tecuci et al., 2011), Disciple-CD (Tecuci et al., 2016a) and Cogent (Tecuci et al., 2015; 2018). MASH also builds on the Disciple multistrategy apprenticeship learning approach (Boicu et al., 2001; Tecuci, 1988; 1998; Tecuci and Hieb, 1996; Tecuci et al., 2000; 2002; 2005; 2007a; 2019; Huang et al., 2020).

## Shared Model of Sense-making

Figure 1 is an overview of a human-machine shared model of sense-making that facilitates the synergistic integration of the analyst's imagination and expertise with the computer's domain knowledge and formal reasoning. It is a general model grounded in the science of evidence (Schum, 2009) and the scientific method of hypothesis generation and testing. Evidence is any observable sign, datum, or item of information that is relevant in deciding whether a hypothesis is true or false (Schum, 2009). The sense-making model consists of three recursive collaborative processes: *Evidence in search of hypotheses; Hypotheses in search of evidence; and Evidentiary assessment of hypotheses.*

The sense-making process starts with an "alerting observation" that may indicate an event of interest. Through *abductive (imaginative) reasoning* which shows that something is *possibly* true, the analyst and MASH generate competing hypotheses that may explain the observation (Peirce, 1955; Eco, 1983; Schum, 2001a; Langley, 2019).

To determine which of these competing hypotheses is true, they use each hypothesis and *deductive reasoning* which shows that something is *necessarily* true, to discover new evidence. The question is: *What evidence would need to be observed if this hypothesis were true?* The reasoning might go as follows: If H were true then the sub-hypotheses $H_1$, $H_2$, and $H_3$ would also need to be true. But if $H_1$ were true then one would need to observe evidence $E_1$, and so on.

A broader question that guides the discovery of evidence is, *What evidence would favor or disfavor hypothesis H?* The decomposition of H is done through a sequence of favoring and disfavoring arguments. These arguments will end in evidence collection requests that will return evidence to test the top hypothesis.

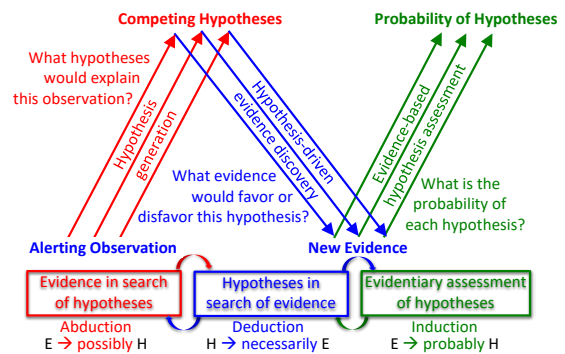

*Figure 1: Human-machine shared model of sense-making.*

Once the evidence is discovered, the analyst and/or MASH use *inductive reasoning* which shows that something is *probably* true, to test each hypothesis. They employ the Wigmorean probabilistic inference network developed during evidence collection, where the probabilities of the bottom hypotheses are assessed based on the collected evidence, and the probabilities of the upper level hypotheses are assessed based on the probabilities of their subhypotheses (Wigmore 1937; Schum, 2001b; Tecuci et al., 2016a). These Wigmorean networks naturally integrate logic and Baconian probability (Cohen, 1977; 1989) with Fuzzy qualifiers (Negoita and Ralescu, 1975; Zadeh, 1983), such as "barely likely," "likely," or "almost certain," being able to deal with all the five characteristics of evidence, namely incompleteness, inconclusiveness, ambiguity, dissonance, and credibility level (Schum, 2001b; Tecuci et al., 2016b, pp. 159-172). This integrated logic and probability system uses the min/max probability combination rules common to the Baconian and the Fuzzy probability views. These rules are much simpler than the Bayesian probability combination rule, which is important for the human understandability of the analysis.

*Evidence in search of hypotheses*, *hypotheses in search of evidence*, and *evidentiary assessment of hypotheses* are collaborative processes that support each other in recursive calls, as shown in the bottom part of Figure 1. For example, the discovery of new evidence may lead to the modification of the existing hypotheses or the generation of new ones that, in turn, lead to the search and discovery of new evidence. Also, inconclusive testing of the hypotheses leads to the need of discovering additional evidence.

## Multi-Agent System Architecture

We have developed MASH as a proof of concept multi-agent system that an analyst can instruct to perform sense-making, as a teacher would instruct a student, through a process that is significantly easier and faster than the typical knowledge engineering approach where the agent is developed by a knowledge engineer who acquires the knowledge from the analyst and encodes it into the agent's knowledge base. Figure 2 shows the overall architecture of MASH, for both its training and its use.

First the analyst demonstrates to the *Mixed-Initiative Learning and Reasoning Assistant* how to determine whether certain activities of interest are taking place. As a result the agent learns general reasoning rules that are stored in the *Knowledge Base*. The *Autonomous Multi-Agent System* uses this knowledge base to automatically reason about other situations as the analyst would. The MASH-generated analysis is reviewed and possibly revised by the analyst. As a result, MASH refines the previously learned rules and learns additional ones, becoming increasingly more competent.

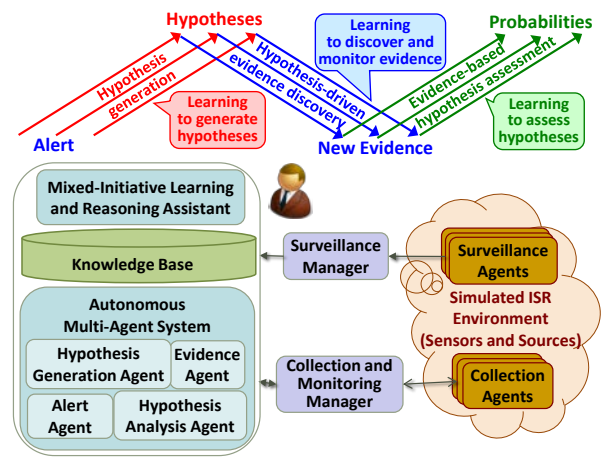

*Figure 2. The overall architecture of MASH.*

The process of teaching and using the system is summarized in Figure 3, and illustrated in the next sections that will show how MASH will be instructed to automatically recognize when a country is producing a certain type of weapon or weapons-related material, such as chemical warfare agents, centrifuge-enriched uranium, or stealth fighter aircraft.

Another important component of the architecture is the *Simulated ISR Environment* that not only enables the testing of automatic sense-making but also facilitates the transition to real data sources and real environments.

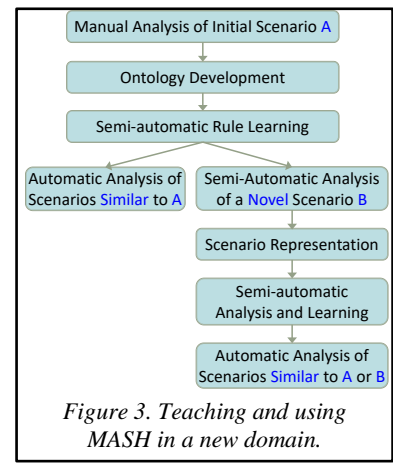

*Figure 3. Teaching and using MASH in a new domain.*

## Demonstration of Sensemaking

Agent instruction starts by demonstrating to MASH how to analyze a specific scenario, such as the one in Table 1.

*Table 1. The Bogustan scenario.*

| |
|---|
| The country Bogustan was building a new chemical plant at Tanan that was nearing completion; the plant's purpose was not known. Bogustan was suspected of harboring weapons of mass destruction ambitions. A reconnaissance asset conducting a routine quarterly overflight detected heat signatures at the Tanan facility on 2/25/2020. |

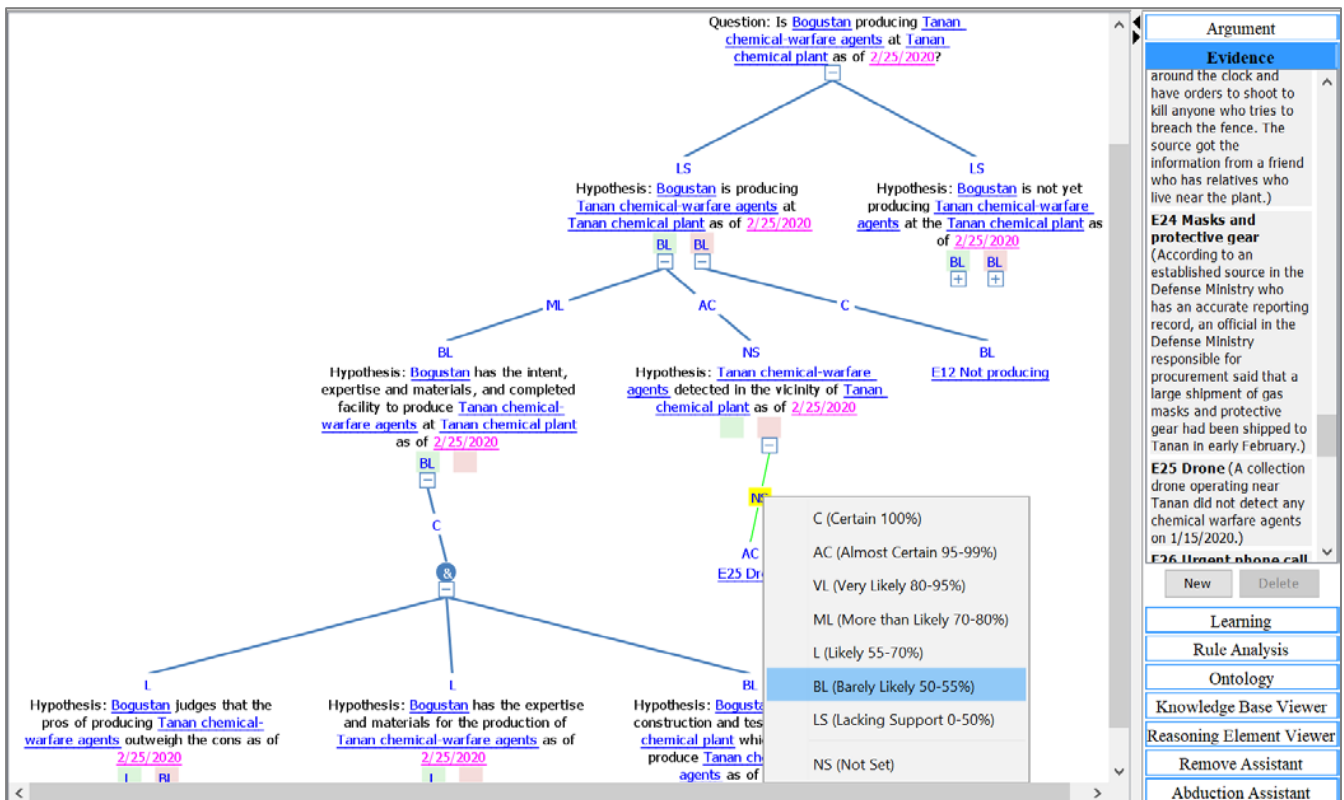

*Figure 4. The system's interface for hypothesis analysis.*

MASH supports the analyst in developing the analysis for answering the question:

> Is Bogustan producing Tanan chemical-warfare agents at Tanan chemical plant as of 2/25/2020?

Figure 4 shows the system's interface for hypothesis analysis. The upper left-hand side pane is the *Whiteboard area* where the analysis is constructed. The upper right-hand side pane is the *Assistants area* that includes several assistants, each helping the user perform a group of related operations. The currently selected one is the *Evidence assistant* that lists the current evidence items. The analyst clicks on an item, then drags and drops it on the relevant hypothesis, either on the green square (if the evidence item favors the truthfulness of the hypothesis) or on the pink square (if it disfavors its truthfulness). The analyst then double-clicks on the NS (Not Set) values of relevance and credibility, and selects the corresponding probability values from the displayed list. For example, the analyst selected BL (Barely Likely, 50-55%) for the relevance of E25 Drone (A collection drone operating near Tanan did not detect any chemical warfare agents on 1/15/2020) because the sensor reported this information a month ago. The information is somewhat dated. More current information would have a higher relevance.

As shown in Figure 4, the analyst considered two competing hypotheses:

> Bogustan is producing Tanan chemical-warfare agents at Tanan chemical plant as of 2/25/2020.
>
> Bogustan is not yet producing Tanan chemical-warfare agents at the Tanan chemical plant as of 2/25/2020.

There are two favoring arguments that would support a conclusion that Bogustan is producing chemical warfare agents:

> Bogustan has the intent to produce chemical warfare agents and the expertise and materials to do so, and the plant at Tanan is complete and was built to produce such agents.
>
> Chemical warfare agents have been detected in the vicinity of the plant.

There is also a disfavoring argument:

> Source reporting stating that this is not the case.

The analysis fragment that is visible in the whiteboard from Figure 4 is a small part of the entire analysis. MASH supports the analyst in developing a *comprehensive*, *defensible*, and *persuasive* analysis by:

- Making reasoned judgments based on all the available information;
- Rigorously considering favoring and disfavoring evidence in the context of large and complex arguments;
- Distinguishing between circumstantial evidence and

- more conclusive (direct) evidence;
- Taking into account the credibility and relevance of evidence;
- Responding to new information without starting over, while minimizing the potential for certain cognitive biases to dismiss or amplify the analytic importance of the new information.

To enable MASH to automatically discover relevant evidence, the analyst needs to insert evidence collection requests under each hypothesis that may have evidence that directly supports that hypothesis. This is done through the simple process illustrated in Figure 5 The analyst right-clicks on the hypothesis "Several areas of the Tanan chemical plant are emitting heat as of 2/25/2020" and selects "Add Collection Task." MASH generates the collection task pattern "Collect evidence from <collection agent> using <function> to determine whether Several areas of the Tanan chemical plant are emitting heat as of 2/25/2020" that the analyst needs to concretize by selecting the collection agent (e.g., thermal imagery sensor) and its collection function (i.e., heat detection).

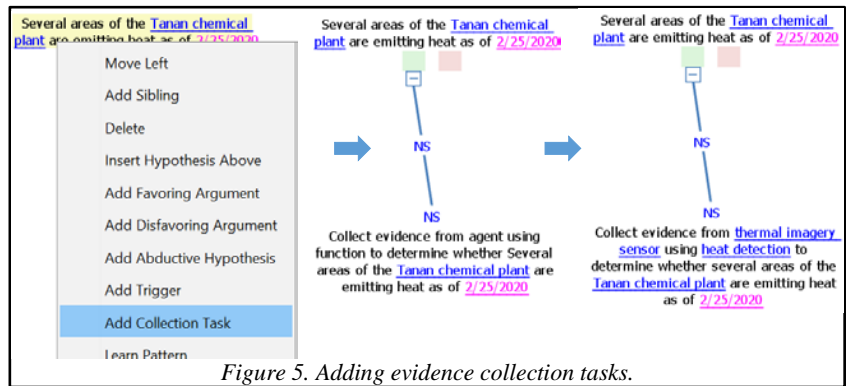

*Figure 5. Adding evidence collection tasks.*

From this analysis MASH learns general rules that will enable it to automatically generate analyses for answering questions of the type

> Is country producing weapons-related product at plant as of date?

in other scenarios, such as

> Is Wokistan producing Wokistan chemical-warfare agents at Bandar chemical plant as of 3/12/2020?

For this, however, it needs an ontology specifying the concepts to which instances (entities) from the question, such as Bogustan, may be generalized (e.g., to country). This has to be done for all the entities appearing in the analysis. The entire ontology for the Bogustan scenario is shown in Figure 6. Notice that it also contains the necessary relationships between the various instances, for example

## Analysis-Driven Ontology Development

The developed analysis of Bogustan answered the question:

> Is Bogustan producing Tanan chemical-warfare agents at Tanan chemical plant as of 2/25/2020?

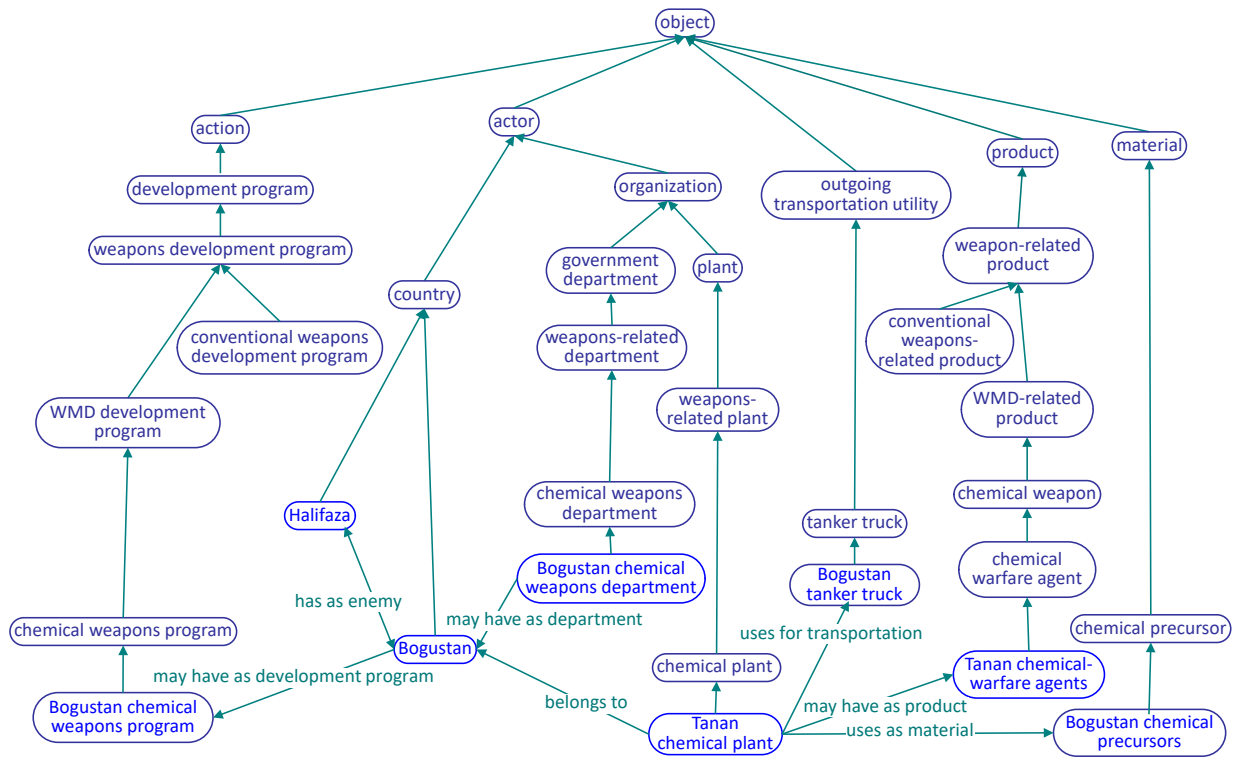

*Figure 6. The ontology for the Bogustan scenario.*

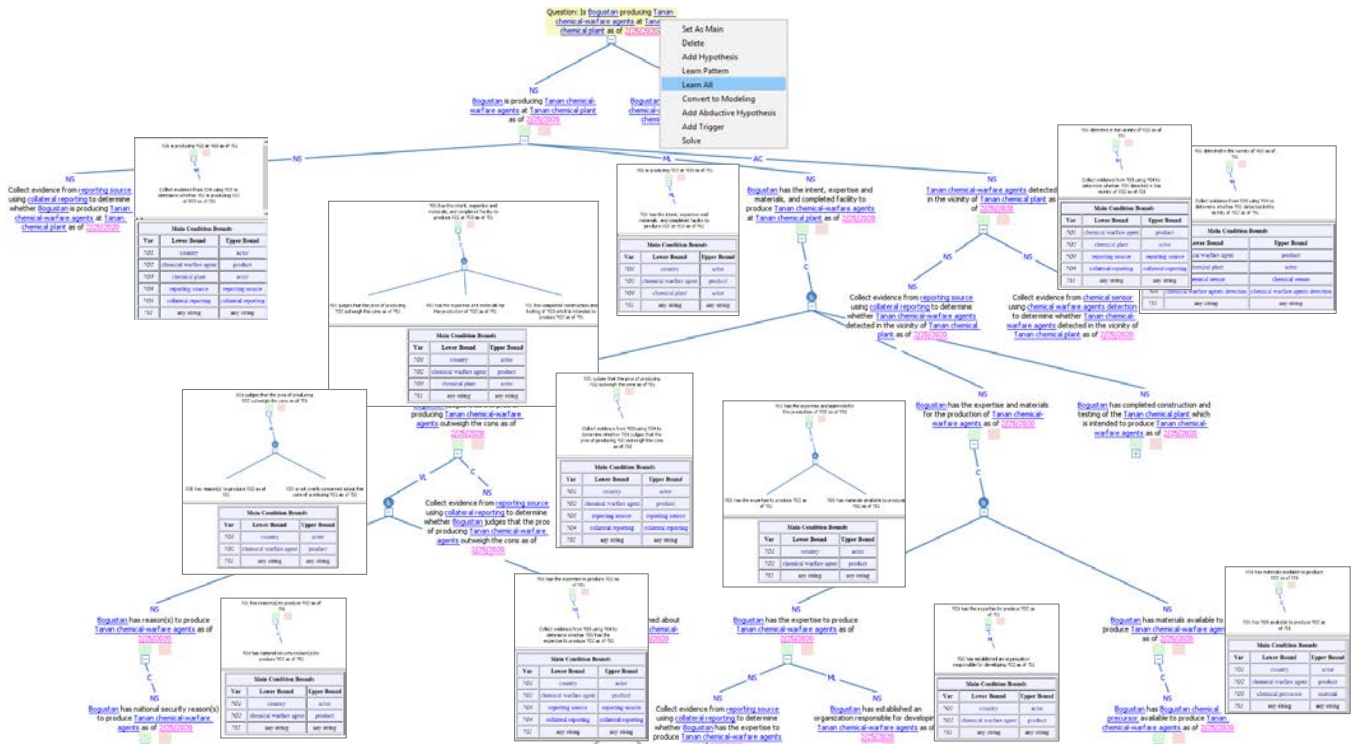

*Figure 7. With a single right-click on the intelligence question, MASH automatically learns all the analysis rules.*

that Tanan chemical plant belongs to Bogustan. *In all, there are only eight instances, seven relationships, and 26 concepts* that are shown in Figure 6.

The ontology is developed using the *Ontology editor*. The ontology language is an extension of RDFS (Allemang et al., 2020; W3C, 2004) with additional features to facilitate learning and evidence representation (Tecuci et a., 2016b).

## Rule Learning

### Automatic Rule Learning

All the 97 analysis rules corresponding to the developed Bogustan analysis are learned with a single click on "Learn All" from the drop-down menu invoked with a right-click on the intelligence question, as shown in Figure 7. Each rule is an ontology-based generalization of an argument for or against a hypothesis.

### Mixed-Initiative Rule Refinement

The analyst uses the *Rule Analysis assistant to* identify the arguments containing new instances in sub-hypotheses whose presence need to be explained by connecting them with instances from the top hypothesis, using mixed-initiative interaction (Tecuci et al., 2007b). For example, the left-hand side of Figure 8 shows such an argument where its sub-hypothesis contains the new instance Halifaza. The learned rule is shown in the middle of Figure 8. Notice that this rule contains the variable ?O3 (generated for Halifaza) in the sub-hypothesis, whose value is not restricted in any way by the values of the variables in the top hypothesis. Therefore, when applying this rule, MASH can instantiate ?O3 with any country or actor.

When the analyst double-clicks on this argument in the *Rule Analysis assistant*, MASH selects the argument in the *Whiteboard* and displays the instance that needs to be explained (Halifaza) together with the possible explanation (Bogustan has as enemy Halifaza) in the *Explanations browser* of the *Learning assistant*, as shown in Figure 9.

The analyst selects the explanation by right-clicking on the feature has as enemy and selecting "Accept." As a result, MASH refines the rule from the middle of Figure 8 as indicated in the right hand side of the figure. Now, MASH will only apply the refined rule when it can instantiate ?O1 with a country or actor that has as enemy the country or actor instantiating ?O3.

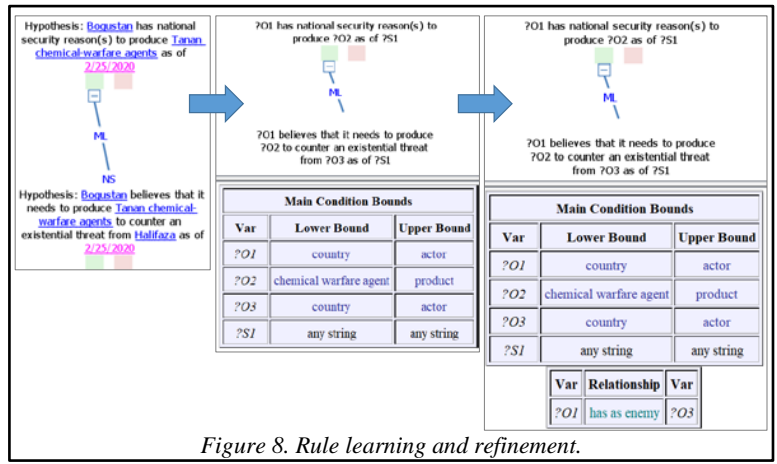

*Figure 8. Rule learning and refinement.*

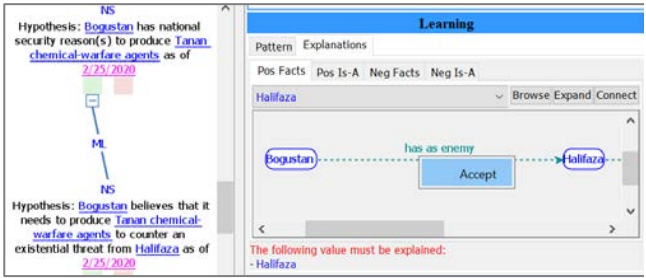

*Figure 9. Mixed-initiative explanation of an argument.*

Only nine out of the 97 learned rules needed to be refined.

**Automatic Analysis of Similar Scenarios**

The rules learned from the Bogustan scenario enable the system to automatically analyze similar scenarios, such as the Wokistan scenario from Table 2.

First the analyst uses the *Ontology editor* to represent the Wokistan scenario. This is similar to the Bogustan scenario in Figure 6, where Bogustan, Tanan, and Halifaza are replaced with Wokistan, Bandar, and Valeria, respectively.

Then, with a single click on "Solve" from the drop-down menu invoked with a right-click on the intelligence question, MASH automatically generates the entire Wokistan analysis in the *Whiteboard* area.

Table 2. The Wokistan scenario.

| The country Wokistan was building a new chemical plant at Bandar that was nearing completion; the plant's purpose was not known. Wokistan was suspected of harboring weapons of mass destruction ambitions. A reconnaissance asset conducting a routine quarterly overflight detected heat signatures at the Bandar facility on 3/12/2020. |
|---|

**Semi-Automatic Analysis of a Novel Scenario**

The rules learned from the Bogustan scenario enable the system to also generate the analysis for a *novel* scenario, such as the Shamland scenario from Table 3 on the production of centrifuge-enriched uranium.

The analyst uses the *Ontology editor* to represent the Shamland scenario shown in Figure 10. This is similar to the Bogustan scenario, where Bogustan, Tanan and Halifaza are replaced with Shamland, Destructville and Agressia. Also, the types of objects are correspondingly updated (e.g., chemical plant is replaced with centrifuge-enriched uranium plant) and Shamland uses two types of critical material inputs, instead of one. Then, with a single click on "Solve" MASH automatically generates the entire Shamland analysis in the *Whiteboard* area.

The analyst browses the automatically generated analysis and can easily modify it where necessary. One update of the Shamland analysis was the addition of an incentive for enriching uranium: economic reasons (electricity shortages and the need of enriched uranium for nuclear power plants). Only 12 new rules where needed to correctly assess the production of centrifuge-enriched uranium.

Table 3. The Shamland scenario.

| The country Shamland was building a large plant at Destructville, whose purpose was not known. Shamland was suspected of wanting to develop nuclear weapons. A reconnaissance asset conducting a routine quarterly overflight detected heat at the Destructville facility on 5/2/2020. |
|---|

**Continued Increase of System's Competence**

With every novel scenario, the system learns a few additional rules, continuously increasing its competence across a broader spectrum of applications. Over time, the system's efficiency also steadily improves because the analyst's effort in addressing new scenarios will correspondingly decrease as the system learns how to reasons about the production of new weapons.

**Semi-Automatic Analysis of Another Novel Scenario**

Table 4 represents the *novel* scenario of a country suspected of producing stealth fighter aircraft. The semantic representation of this scenario is similar to that in Figure 6, where Bogustan, Tanan, and Halifaza are replaced with Violenta, Bemoana, and Smoldera, respectively. Also, the types of involved entities are correspondingly updated (e.g., chemical plant is replaced with aircraft plant) and Violenta does not use any transportation utility because a produced aircraft can move itself.

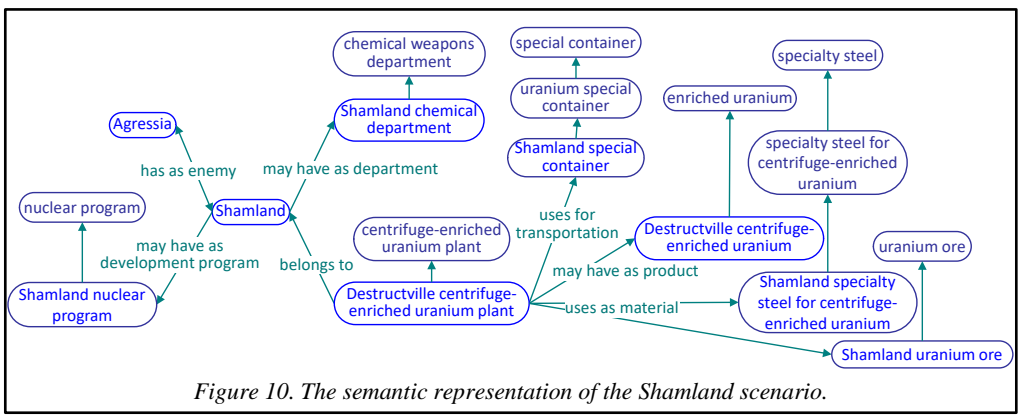

*Figure 10. The semantic representation of the Shamland scenario.*

Table 4. The Violenta scenario.

> During the past four decades, several conflicts have occurred between Violenta and Smoldera in which air power has played an increasingly important role. In the last conflict two years ago, Smoldera's air force, after destroying Violenta's air force, extensively bombed Violenta's armored units and forced Violenta to make major border concessions. On 6/10/2020 several areas of a new plant that Violenta was building at Bemoana were emitting heat signatures. Is Violenta producing stealth fighter aircraft at Bemoana?

Then, with a single click on "Solve" MASH automatically generates the entire Violenta analysis in the *Whiteboard* area. The analyst browses the automatically generated analysis and can modify it where necessary. One update was the addition of two new areas of expertise needed to manufacture stealth aircraft. Only 20 new rules where needed to instruct MASH to assess the production of stealth fighter aircraft.

**Automatic Analysis of a Novel Scenario**

Table 5 represents the novel scenario of a country suspected of producing long-range stealth bomber aircraft.

*Table 5. The Malicia scenario.*

> Goodlanda and Malicia are nuclear-weapon powers who vie for global influence and see each other as arch enemies. Both have long-range strategic bombers that can travel the 6,000-mile distance that separates Goodlanda and Malicia. The bombers, however, are easily detected by long-range radars positioned along the periphery of both countries that provide ample early warning of a possible attack. On 7/15/2020 several areas of a new plant that Malicia was building at Tirinta were emitting heat. Is Malicia producing long-range stealth bombers at Tirinta?

The semantic representation of this scenario is similar to the Violenta scenario, where Violenta, Bemoana, Smoldera, and stealth fighter aircraft, are replaced with Malicia, Tirinta, and Goodlanda, and long-range stealth bomber aircraft respectively.

Then, with a single click on "Solve" MASH automatically generates the entire Malicia analysis in the *Whiteboard* area. *This analysis was complete and correct, no adaptation being necessary.*

## Cognitive Augmentation

As discussed in the previous sections, by using MASH the analyst follows a systematic analysis process that synergistically integrates the user's imaginative reasoning and expertise with the agent's formal reasoning and learned expertise. For example, the analyst imagines the questions to ask and hypothesizes possible answers. MASH helps with developing the arguments by reusing previously learned rules, and guides the evidence collection. The jointly-developed analysis makes very clear the logic, what evidence was used and how, what is not known, and what assumptions have been made. It can be shared with other users, subjected to critical review, and correspondingly improved. As a result, this systematic process leads to the development of defensible and persuasive conclusions. MASH also enables rapid analysis, not only through the reuse of patterns, but also through a drill-down process where a hypothesis may be decomposed to different levels of detail, depending on the available time. It facilitates the analysis of what-if scenarios, where the user may make various assumptions and the assistant automatically determines their influence on the analytic conclusion. The assistant also makes possible the rapid updating of the analysis based on new (or revised) evidence from monitored sources, and assumptions.

## Trust in the Machine-Generated Analysis

Machine-generated analyses are similar to the ones developed by the analyst. The transparency and defensibility of the developed analysis facilitate its review by the analyst. The probabilistic assessments are based on the simple min-max rules common to the Fuzzy and Baconian systems that are intuitive and easy to understand. Moreover, the analyst may review each argument from the analysis, may accept, may revise, or may even reject it. MASH uses the analyst's feedback to further improve the previously learned rules and to learn new ones, to replicate better and better the reasoning of the analyst, and thus insuring increased trust in the generated analyses.

## Conclusions and Future Research

This paper showed how a shared model of sense-making facilitated the synergistic integration of the analyst's imagination and expertise with the computer's knowledge and critical reasoning.

But there are more opportunities for human-machine collaboration with respect to this model, including: mixed-initiative ontology learning, fully-automatic rule learning, deep sense-making through iterative (multi-step) abduction, deduction and induction; advanced analytics (e.g., detection and mitigation of cognitive biases, automatic identification of key evidence and assumptions); natural language interaction, and automatic evidence collection using edge processing with convolutional neural networks.